\documentclass[11pt]{article}

\usepackage[final]{acl}
\usepackage{times}
\usepackage{latexsym}
\usepackage[T1]{fontenc}
\usepackage[utf8]{inputenc}
\usepackage{microtype}
\usepackage{inconsolata}
\usepackage{graphicx}

\usepackage{amsmath,amssymb}
\usepackage{booktabs}
\usepackage{multirow}
\usepackage{array}
\usepackage{algorithm}
\usepackage{algorithmic}
\usepackage{tikz}
\usetikzlibrary{positioning,fit,backgrounds,calc}
\usepackage[table,dvipsnames]{xcolor}
\usepackage{colortbl}
\usepackage{subcaption}

\definecolor{accBlue}{RGB}{79,70,229}
\definecolor{robOrange}{RGB}{220,80,10}
\definecolor{logGreen}{RGB}{22,155,70}
\definecolor{hovAmber}{RGB}{170,120,0}
\definecolor{stageDark}{RGB}{55,55,55}
\definecolor{stageMid}{RGB}{110,110,110}
\definecolor{stageLight}{RGB}{165,165,165}

\newcommand{\gtag}[2]{%
  \colorbox{#1!12}{\textcolor{#1!85!black}{\textbf{\footnotesize #2}}}}

\title{ComplianceAtlas: A Taxonomy-Driven Agentic AI Framework for Automated AI Act Gap Analysis and Remediation}

\author{Ilias Driouich \quad Hongliu Cao \quad Eoin Thomas  \quad Yannick Devaux  \quad Agnes Brodelle  \\
  Amadeus SAS\\ 
  \texttt{\{ilias.driouich\}@amadeus.com}}

\begin{document}
\maketitle

\begin{abstract}
Assessing regulatory compliance for high-risk AI systems requires systematic
identification of technical gaps such as missing controls, inadequate processes, or
 shallow evaluation of practices. Compliance officers
typically administer structured questionnaires collecting evidence of current
practices, where organizations respond with descriptions of their
existing processes, technical capabilities, and architectural choices. However,
translating these compliance questionnaire responses into typed,
actionable gap diagnoses remains unsystematic. In this paper, we present a
literature-grounded taxonomy of technical compliance gaps
covering areas such as accuracy, robustness, logging, and human oversight, and embed it in
\textbf{GapToCompliance}, a two-stage agentic framework coordinating two
specialized agents: a \emph{Gap Extraction Agent} that performs hierarchical
reasoning from compliance areas to specific gap types, and a \emph{Remediation Planning Agent} that ranks remediation
actions from a remediation catalog via multi-signal scoring combining semantic similarity, regulatory
alignment, and effort estimation. We evaluate this framework against four
automatic compliance gap detection baselines across 9 AI systems and 6 AI families. Our approach achieves a gap-type F1 of 0.831 (vs.\ 0.533 for the strongest baseline) and
remediation P@5 of 0.872 (vs.\ 0.635 for retrieval methods), demonstrating
promising potential for automated compliance workflows. The structured agent
coordination reveals a path toward human-in-the-loop compliance systems where
automated gap extraction reduces expert burden while maintaining auditability,
and the taxonomy-driven design enables extension to emerging regulatory
frameworks beyond the EU AI Act.

\end{abstract}

\section{Introduction}
\label{sec:intro}

The EU AI Act~\cite{eu_ai_act_2024} introduces binding obligations for providers
of high risk AI systems across areas such as accuracy, robustness, logging, and
human oversight. Organizations commonly operationalize these obligations through
structured compliance questionnaires that transform regulatory requirements into
observable assessment items.

Although these questionnaires are useful for identifying missing controls or
incomplete practices, they rarely explain the precise technical nature of the
problem. Two systems may receive similar questionnaire answers while suffering
from fundamentally different operational deficiencies. One system may lack
formal evaluation metrics, another may evaluate correctly but lack production
monitoring, while a third may monitor runtime behavior without defining
acceptable thresholds. These situations require different remediation
strategies, technical expertise, and implementation effort. This distinction is important because modern ML systems fail in many
operationally distinct ways. Previous work on ML production systems has
highlighted issues such as monitoring gaps, training serving skew, concept
drift, fairness instability, and insufficient observability
\cite{sculley2015debt,breck2017mltestscore,ashmore2021assuring}. Existing
compliance workflows frequently capture that a requirement is unsatisfied
without identifying the underlying failure category.

Current automated approaches also remain limited. Checklist based systems focus
on whether a requirement is present but do not characterize the technical nature
of the deficiency. Retrieval based systems rely heavily on semantic similarity,
which often causes them to recommend remediation actions that appear
linguistically related but are technically inappropriate. Behavioral testing
approaches evaluate deployed systems but usually do not generate structured
remediation guidance. To address these limitations, we introduce
\textbf{GapToCompliance}, a taxonomy driven framework for automated compliance
gap extraction and remediation recommendation under the EU AI Act.

The framework is built around two main contributions. First, we develop a literature grounded taxonomy of technical compliance gaps
covering four major requirement areas: accuracy, robustness, logging, and human
oversight. The taxonomy defines typed operational deficiencies together with
their detection signals, applicability constraints, and remediation
characteristics. Second, we integrate this taxonomy into a two stage agentic pipeline. The
Gap Extraction Agent combines deterministic rules, constrained LLM reasoning,
and cross question reconciliation to identify typed compliance gaps from
questionnaire evidence. The Remediation Planning Agent ranks remediation actions
through a scoring function combining semantic relevance, regulatory alignment,
and AI family compatibility.

The framework is evaluated on 41 questionnaire items collected from 9 AI
systems spanning multiple AI families and requirement areas. Experimental
results show that GapToCompliance achieves strong improvements over retrieval
based and few shot baselines, particularly in situations requiring structured
regulatory reasoning and architectural context awareness.

The results demonstrate that explicit gap typing is essential for reliable
compliance automation because semantically similar deficiencies frequently
require substantially different remediation strategies.

\section{Related Work}
\label{sec:related}

\paragraph{ML engineering and operational reliability.}
Research on ML production systems has highlighted the complexity of maintaining
reliable and observable AI systems in operational environments.
\citeauthor{sculley2015debt}~\citep{sculley2015debt} introduced the concept
of hidden technical debt in ML systems, identifying recurring issues such as
training serving skew, monitoring deficiencies, and unstable evaluation
pipelines. \citeauthor{breck2017mltestscore}~\citep{breck2017mltestscore}
proposed operational readiness criteria for production ML systems spanning data
quality, infrastructure robustness, and runtime monitoring.
\citeauthor{ashmore2021assuring}~\citep{ashmore2021assuring} further
extended lifecycle assurance principles for AI systems across development and
deployment phases. These works establish the operational foundations for several taxonomy
categories introduced in this paper, particularly gaps related to monitoring,
evaluation, drift detection, and production readiness.

\paragraph{Robustness, fairness, and uncertainty estimation.}
Fairness and robustness assessment have become central topics in trustworthy AI
research. \citeauthor{mitchell2019modelcards}
\citep{mitchell2019modelcards} emphasized disaggregated evaluation and
structured reporting practices through model cards.
\citeauthor{barocas2019fairness}~\citep{barocas2019fairness} and
\citeauthor{mehrabi2021survey}~\citep{mehrabi2021survey} surveyed fairness
definitions and mitigation approaches across ML systems.

Adversarial robustness research by
\citeauthor{goodfellow2015explaining}
\citep{goodfellow2015explaining} and
\citeauthor{madry2018towards}~\citep{madry2018towards} established common
evaluation techniques for vulnerability assessment.
Uncertainty estimation approaches proposed by
\citeauthor{lakshminarayanan2017simple}
\citep{lakshminarayanan2017simple} and
\citeauthor{ovadia2019trust}~\citep{ovadia2019trust} motivated the
robustness taxonomy categories related to predictive confidence and reliability.

\paragraph{Human oversight and human AI interaction.}
Human oversight has been widely studied in the context of automation and
decision support systems. \citeauthor{parasuraman2000model}
\citep{parasuraman2000model} proposed foundational models for levels of
automation and human supervision.
\citeauthor{amershi2019guidelines}
\citep{amershi2019guidelines} introduced practical human AI interaction
guidelines focusing on interpretability, intervention capabilities, and user
feedback mechanisms.
\citeauthor{cummings2004automation}
\citep{cummings2004automation} and
\citeauthor{bucinca2021trust}~\citep{bucinca2021trust} studied trust,
notification design, and intervention effectiveness in AI assisted systems.

These studies provide the conceptual basis for the human oversight categories
defined in our taxonomy.

\paragraph{AI governance and compliance automation.}
Prior work on AI governance has proposed structured assessment methodologies for
trustworthy AI systems.
\citeauthor{raji2020closing}~\citep{raji2020closing} introduced internal AI
auditing methodologies for accountability and traceability.
The ALTAI framework~\cite{hleg2020altai} and the NIST AI RMF
\cite{nist2023airmf} formalized governance oriented assessment structures for
trustworthy AI evaluation.

Automated compliance analysis has also been explored in NLP and retrieval based
systems. \citeauthor{sleimi2018automated}~\citep{sleimi2018automated} studied
automated legal metadata extraction and requirement checking using NLP pipelines.
Retrieval augmented generation approaches
\cite{lewis2020retrieval,gao2023retrieval} improved contextual reasoning in
knowledge intensive tasks but remain vulnerable to retrieval ambiguity and
semantic overlap.
\citeauthor{wu2024ragchecker}~\citep{wu2024ragchecker} specifically
highlighted retrieval precision failures in RAG systems.

Unlike previous work, our approach introduces a typed taxonomy of technical
compliance deficiencies grounded simultaneously in AI governance requirements
and operational ML engineering literature. The proposed framework also links
gap typing directly to remediation ranking, enabling more reliable and
traceable compliance planning.

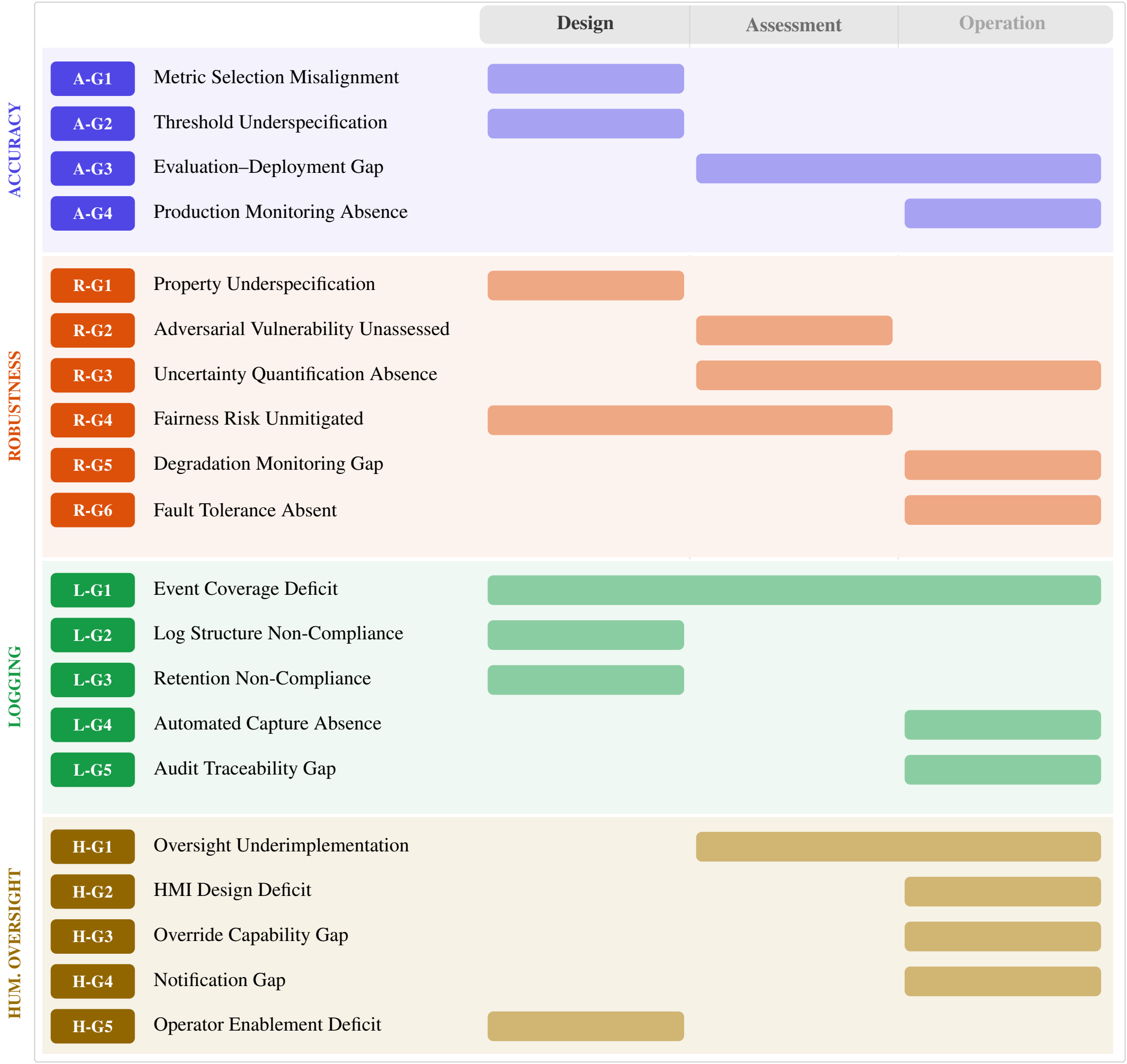
\begin{figure*}[t]
\centering
\begin{tikzpicture}[font=\scriptsize, x=1cm, y=1cm]

\colorlet{accC}{accBlue}
\colorlet{robC}{robOrange}
\colorlet{logC}{logGreen}
\colorlet{hovC}{hovAmber}


\fill[stageDark!12, rounded corners=3pt]
     (5.40, 0.10) rectangle (13.3, 0.58);
\draw[stageDark!20, thin] (8.02,  0.10) -- (8.02,  0.58);
\draw[stageDark!20, thin] (10.62, 0.10) -- (10.62, 0.58);
\node[font=\scriptsize\bfseries, text=stageDark]  at (6.72,  0.34) {Design};
\node[font=\scriptsize\bfseries, text=stageMid]   at (9.32,  0.34) {Assessment};
\node[font=\scriptsize\bfseries, text=stageLight] at (11.92, 0.34) {Operation};
\node[font=\tiny,  text=gray, anchor=west]        at (13.30, 0.34) {};

\draw[gray!22, densely dashed, thin] (8.02,  0.10) -- (8.02,  -11.62);
\draw[gray!22, densely dashed, thin] (10.62, 0.10) -- (10.62, -11.62);

\fill[accC!7] (-0.05, 0.05) rectangle (13.3, -2.50);
\node[font=\tiny\bfseries, text=accC, rotate=90] at (-0.40, -1.22) {ACCURACY};
\node[font=\scriptsize\bfseries, text=accC]       at (14.00, -1.22) {};

\fill[accC, rounded corners=2pt] (0.05,-0.12) rectangle (1.10,-0.55);
\node[white,font=\tiny\bfseries] at (0.575,-0.335) {A-G1};
\node[anchor=west, font=\scriptsize] at (1.20,-0.335) {Metric Selection Misalignment};
\fill[accC!50, rounded corners=2pt] (5.50,-0.15) rectangle (7.95,-0.52);
\node[font=\tiny, text=gray, anchor=west] at (13.30,-0.335) {};

\fill[accC, rounded corners=2pt] (0.05,-0.68) rectangle (1.10,-1.11);
\node[white,font=\tiny\bfseries] at (0.575,-0.895) {A-G2};
\node[anchor=west, font=\scriptsize] at (1.20,-0.895) {Threshold Underspecification};
\fill[accC!50, rounded corners=2pt] (5.50,-0.71) rectangle (7.95,-1.08);
\node[font=\tiny, text=gray, anchor=west] at (13.30,-0.895) {};

\fill[accC, rounded corners=2pt] (0.05,-1.24) rectangle (1.10,-1.67);
\node[white,font=\tiny\bfseries] at (0.575,-1.455) {A-G3};
\node[anchor=west, font=\scriptsize] at (1.20,-1.455) {Evaluation--Deployment Gap};
\fill[accC!50, rounded corners=2pt] (8.10,-1.27) rectangle (13.15,-1.64);
\node[font=\tiny, text=gray, anchor=west] at (13.30,-1.455) {};

\fill[accC, rounded corners=2pt] (0.05,-1.80) rectangle (1.10,-2.23);
\node[white,font=\tiny\bfseries] at (0.575,-2.015) {A-G4};
\node[anchor=west, font=\scriptsize] at (1.20,-2.015) {Production Monitoring Absence};
\fill[accC!50, rounded corners=2pt] (10.70,-1.83) rectangle (13.15,-2.20);
\node[font=\tiny, text=gray, anchor=west] at (13.30,-2.015) {};

\fill[robC!7] (-0.05,-2.55) rectangle (13.3,-6.30);
\node[font=\tiny\bfseries, text=robC, rotate=90] at (-0.40,-4.42) {ROBUSTNESS};
\node[font=\scriptsize\bfseries, text=robC]       at (14.00,-4.42) {};

\fill[robC, rounded corners=2pt] (0.05,-2.70) rectangle (1.10,-3.13);
\node[white,font=\tiny\bfseries] at (0.575,-2.915) {R-G1};
\node[anchor=west, font=\scriptsize] at (1.20,-2.915) {Property Underspecification};
\fill[robC!50, rounded corners=2pt] (5.50,-2.73) rectangle (7.95,-3.10);
\node[font=\tiny, text=gray, anchor=west] at (13.30,-2.915) {};

\fill[robC, rounded corners=2pt] (0.05,-3.26) rectangle (1.10,-3.69);
\node[white,font=\tiny\bfseries] at (0.575,-3.475) {R-G2};
\node[anchor=west, font=\scriptsize] at (1.20,-3.475) {Adversarial Vulnerability Unassessed};
\fill[robC!50, rounded corners=2pt] (8.10,-3.29) rectangle (10.55,-3.66);
\node[font=\tiny, text=gray, anchor=west] at (13.30,-3.475) {};

\fill[robC, rounded corners=2pt] (0.05,-3.82) rectangle (1.10,-4.25);
\node[white,font=\tiny\bfseries] at (0.575,-4.035) {R-G3};
\node[anchor=west, font=\scriptsize] at (1.20,-4.035) {Uncertainty Quantification Absence};
\fill[robC!50, rounded corners=2pt] (8.10,-3.85) rectangle (13.15,-4.22);
\node[font=\tiny, text=gray, anchor=west] at (13.30,-4.035) {};

\fill[robC, rounded corners=2pt] (0.05,-4.38) rectangle (1.10,-4.81);
\node[white,font=\tiny\bfseries] at (0.575,-4.595) {R-G4};
\node[anchor=west, font=\scriptsize] at (1.20,-4.595) {Fairness Risk Unmitigated};
\fill[robC!50, rounded corners=2pt] (5.50,-4.41) rectangle (10.55,-4.78);
\node[font=\tiny, text=gray, anchor=west] at (13.30,-4.595) {};

\fill[robC, rounded corners=2pt] (0.05,-4.94) rectangle (1.10,-5.37);
\node[white,font=\tiny\bfseries] at (0.575,-5.155) {R-G5};
\node[anchor=west, font=\scriptsize] at (1.20,-5.155) {Degradation Monitoring Gap};
\fill[robC!50, rounded corners=2pt] (10.70,-4.97) rectangle (13.15,-5.34);
\node[font=\tiny, text=gray, anchor=west] at (13.30,-5.155) {};

\fill[robC, rounded corners=2pt] (0.05,-5.50) rectangle (1.10,-5.93);
\node[white,font=\tiny\bfseries] at (0.575,-5.715) {R-G6};
\node[anchor=west, font=\scriptsize] at (1.20,-5.715) {Fault Tolerance Absent};
\fill[robC!50, rounded corners=2pt] (10.70,-5.53) rectangle (13.15,-5.90);
\node[font=\tiny, text=gray, anchor=west] at (13.30,-5.715) {};

\fill[logC!7] (-0.05,-6.35) rectangle (13.3,-9.50);
\node[font=\tiny\bfseries, text=logC, rotate=90] at (-0.40,-7.92) {LOGGING};
\node[font=\scriptsize\bfseries, text=logC]       at (14.00,-7.92) {};

\fill[logC, rounded corners=2pt] (0.05,-6.50) rectangle (1.10,-6.93);
\node[white,font=\tiny\bfseries] at (0.575,-6.715) {L-G1};
\node[anchor=west, font=\scriptsize] at (1.20,-6.715) {Event Coverage Deficit};
\fill[logC!50, rounded corners=2pt] (5.50,-6.53) rectangle (13.15,-6.90);
\node[font=\tiny, text=gray, anchor=west] at (13.30,-6.715) {};

\fill[logC, rounded corners=2pt] (0.05,-7.06) rectangle (1.10,-7.49);
\node[white,font=\tiny\bfseries] at (0.575,-7.275) {L-G2};
\node[anchor=west, font=\scriptsize] at (1.20,-7.275) {Log Structure Non-Compliance};
\fill[logC!50, rounded corners=2pt] (5.50,-7.09) rectangle (7.95,-7.46);
\node[font=\tiny, text=gray, anchor=west] at (13.30,-7.275) {};

\fill[logC, rounded corners=2pt] (0.05,-7.62) rectangle (1.10,-8.05);
\node[white,font=\tiny\bfseries] at (0.575,-7.835) {L-G3};
\node[anchor=west, font=\scriptsize] at (1.20,-7.835) {Retention Non-Compliance};
\fill[logC!50, rounded corners=2pt] (5.50,-7.65) rectangle (7.95,-8.02);
\node[font=\tiny, text=gray, anchor=west] at (13.30,-7.835) {};

\fill[logC, rounded corners=2pt] (0.05,-8.18) rectangle (1.10,-8.61);
\node[white,font=\tiny\bfseries] at (0.575,-8.395) {L-G4};
\node[anchor=west, font=\scriptsize] at (1.20,-8.395) {Automated Capture Absence};
\fill[logC!50, rounded corners=2pt] (10.70,-8.21) rectangle (13.15,-8.58);
\node[font=\tiny, text=gray, anchor=west] at (13.30,-8.395) {};

\fill[logC, rounded corners=2pt] (0.05,-8.74) rectangle (1.10,-9.17);
\node[white,font=\tiny\bfseries] at (0.575,-8.955) {L-G5};
\node[anchor=west, font=\scriptsize] at (1.20,-8.955) {Audit Traceability Gap};
\fill[logC!50, rounded corners=2pt] (10.70,-8.77) rectangle (13.15,-9.14);
\node[font=\tiny, text=gray, anchor=west] at (13.30,-8.955) {};

\fill[hovC!10] (-0.05,-9.55) rectangle (13.3,-12.50);
\node[font=\tiny\bfseries, text=hovC!90!black, rotate=90] at (-0.40,-11.12) {HUM.\ OVERSIGHT};
\node[font=\scriptsize\bfseries, text=hovC!90!black]       at (14.00,-11.12) {};

\fill[hovC!85!black, rounded corners=2pt] (0.05,-9.70) rectangle (1.10,-10.13);
\node[white,font=\tiny\bfseries] at (0.575,-9.915) {H-G1};
\node[anchor=west, font=\scriptsize] at (1.20,-9.915) {Oversight Underimplementation};
\fill[hovC!55, rounded corners=2pt] (8.10,-9.73) rectangle (13.15,-10.10);
\node[font=\tiny, text=gray, anchor=west] at (13.30,-9.915) {};

\fill[hovC!85!black, rounded corners=2pt] (0.05,-10.26) rectangle (1.10,-10.69);
\node[white,font=\tiny\bfseries] at (0.575,-10.475) {H-G2};
\node[anchor=west, font=\scriptsize] at (1.20,-10.475) {HMI Design Deficit};
\fill[hovC!55, rounded corners=2pt] (10.70,-10.29) rectangle (13.15,-10.66);
\node[font=\tiny, text=gray, anchor=west] at (13.30,-10.475) {};

\fill[hovC!85!black, rounded corners=2pt] (0.05,-10.82) rectangle (1.10,-11.25);
\node[white,font=\tiny\bfseries] at (0.575,-11.035) {H-G3};
\node[anchor=west, font=\scriptsize] at (1.20,-11.035) {Override Capability Gap};
\fill[hovC!55, rounded corners=2pt] (10.70,-10.85) rectangle (13.15,-11.22);
\node[font=\tiny, text=gray, anchor=west] at (13.30,-11.035) {};

\fill[hovC!85!black, rounded corners=2pt] (0.05,-11.38) rectangle (1.10,-11.81);
\node[white,font=\tiny\bfseries] at (0.575,-11.595) {H-G4};
\node[anchor=west, font=\scriptsize] at (1.20,-11.595) {Notification Gap};
\fill[hovC!55, rounded corners=2pt] (10.70,-11.41) rectangle (13.15,-11.78);
\node[font=\tiny, text=gray, anchor=west] at (13.30,-11.595) {};

\fill[hovC!85!black, rounded corners=2pt] (0.05,-11.94) rectangle (1.10,-12.37);
\node[white,font=\tiny\bfseries] at (0.575,-12.155) {H-G5};
\node[anchor=west, font=\scriptsize] at (1.20,-12.155) {Operator Enablement Deficit};
\fill[hovC!55, rounded corners=2pt] (5.50,-11.97) rectangle (7.95,-12.34);
\node[font=\tiny, text=gray, anchor=west] at (13.30,-12.155) {};

\draw[gray!35, thin, rounded corners=1pt]
     (-0.15, 0.62) rectangle (13.45, -12.60);

\end{tikzpicture}
\caption{\textbf{GapToCompliance Taxonomy of Technical Compliance Gaps.}
  Coloured stage bands indicate when each gap type typically manifests across
  the compliance lifecycle. A bar spanning multiple stages signifies that the
  underlying deficiency can emerge or propagate across those phases.
  Effort ratings reflect the remediation complexity modulated by AI family.}
\label{fig:taxonomy}
\end{figure*}


\section{A Literature-Grounded Taxonomy of Technical Compliance Gaps}
\label{sec:taxonomy}

\subsection{Derivation Methodology}

The taxonomy was derived in three stages. In the literature anchoring
stage, we reviewed ML engineering, fairness, robustness, and human-AI
interaction research to enumerate distinct ways technical requirements fail in
practice. In the regulatory mapping stage, we mapped each failure mode
to the specific AI Act article and questionnaire item that makes it a compliance
obligation. In the signal grounding stage, we validated each type against
real questionnaire instances from six AI families, confirming that: (i)~the
type appears in practice; (ii)~it is detectable from the combination of answer
value and evidence note via typed signals; and (iii)~its remediation is
qualitatively distinct from all other types.

Each gap type is defined by five components: compliance semantics
anchored in literature and regulation; absence signals (keywords that
trigger the gap); negation signals (keywords that suppress it); a
remediation orientation with family-modulated effort; and cross-family applicability constraints.

The implementation effort is sized according to the expected implementation burden.
\emph{Low} effort corresponds to documentation updates, policy revisions, and
configuration changes; \emph{Medium} effort corresponds to process changes,
tooling integration, and dedicated test suites; and \emph{High} effort
corresponds to architectural changes, new infrastructure, or model retraining.
This effort label is used both to characterize the taxonomy and to guide
remediation prioritization in the planning stage.

\noindent
Figure~\ref{fig:taxonomy} presents the full taxonomy in lifecycle context.
Three structural observations bear on detection design. First, accuracy and
logging gaps cluster in the Design column: they reflect decisions made before a
system is deployed (metric selection, threshold specification, log schema),
and deterministic rule matching against questionnaire evidence is sufficient for
most instances. Second, robustness and oversight gaps are predominantly
Assessment- or Operation-stage: adversarial testing, drift monitoring, and HMI
adequacy can only be evaluated against a running or tested system, motivating
the LLM fallback for ambiguous evidence. Third, multi-stage bars (A-G3, R-G3,
R-G4, H-G1, L-G1) reflect deficiencies that span the lifecycle and are the
principal source of cross-question reconciliation challenges.

\subsection{Accuracy Gaps}
\label{sec:taxonomy_acc}

\textbf{A-G1: Metric Selection Misalignment.}
No accuracy metrics are defined, or the metrics used are inappropriate for the
AI family (e.g., RMSE applied to a generative
system)~\cite{breck2017mltestscore,mitchell2019modelcards}.
Anchor: Art.~9(7), Art.~15.

\textbf{A-G2: Threshold Underspecification.}
Metrics are measured but no formal absolute acceptance thresholds are defined.
Relative-only descriptions (``better than baseline'') or thresholds scoped only
to the model layer all trigger this type~\cite{breck2017mltestscore}.
Anchor: Art.~9(7). 

\textbf{A-G3: Evaluation--Deployment Gap.}
Evaluation is performed only under offline or laboratory conditions; applies
exclusively to discriminative families (classification, regression, ranking,
vision)~\cite{sculley2015debt,paleyes2022challenges}. Anchor: Art.~9(6)(a).

\textbf{A-G4: Production Monitoring Absence.}
A rigorous offline pipeline exists but no live monitoring mechanism (dashboard,
alert, or automated comparison) is deployed. ~\cite{breck2017mltestscore,klaise2021alibi}. Anchor: Art.~9(1)(f), Art.~72.

\subsection{Robustness Gaps}
\label{sec:taxonomy_rob}

\textbf{R-G1--R-G2.}
R-G1 flags the absence of formal robustness property definitions
(reliability, stability, sensitivity, reachability)~\cite{seshia2018formal};
R-G2 flags the absence of adversarial testing, including prompt injection detection for
LLMs~\cite{goodfellow2015explaining,madry2018towards,carlini2019evaluating}.

\textbf{R-G3--R-G4.}
R-G3 detects the absence of calibrated uncertainty outputs or, for RAG systems,
missing retrieval confidence
thresholds~\cite{lakshminarayanan2017simple,ovadia2019trust}.
R-G4 captures unmitigated fairness risk: potential bias sources identified but
lacking formal mitigation or
disclosure~\cite{mehrabi2021survey,barocas2019fairness,chouldechova2017fair}.

\textbf{R-G5--R-G6.}
R-G5 detects the absence of continuous production monitoring for model or data
drift~\cite{gama2014concept,klaise2021alibi}; R-G6 detects the absence of
redundancy, fail-safe, rollback, or graceful degradation
mechanisms~\cite{avizienis2004basic}. 

\subsection{Logging Gaps}
\label{sec:taxonomy_log}

L-G1 (\emph{Event Coverage Deficit}) flags incomplete capture of mandatory
event categories (Art.~12, Art.~19). L-G2 (\emph{Log Structure Non-Compliance})
detects missing mandatory fields (AI system ID, timestamp, input reference,
output, operator ID, risk flags) or broken linkage to input-output pairs. L-G3 (\emph{Retention Non-Compliance}) applies a
deterministic comparison: stated retention period vs.\ the 6-month minimum of
Art.~19. L-G4
(\emph{Automated Capture Absence}) flags manual or sampling-only logging. L-G5
(\emph{Audit Traceability Gap}) detects logs not linked to AI system version
information in a way that enables reproducibility~\cite{raji2020closing}.

\subsection{Human Oversight Gaps}
\label{sec:taxonomy_ho}

H-G1 (\emph{Oversight Underimplementation}) fires when identified oversight
measures are absent or only partially deployed~\cite{parasuraman2000model};
H-G2 (\emph{HMI Design Deficit}) when no interface enables designated persons
to understand system behaviour and
intervene~\cite{amershi2019guidelines,endsley1995toward};
H-G3 (\emph{Override Capability Gap}) when the three intervention modalities
(disregard, override, reverse) are not independently satisfied within
operational response time~\cite{bucinca2021trust};
H-G4 (\emph{Notification Gap}) when no automated notification mechanism
exists~\cite{cummings2004automation};
H-G5 (\emph{Operator Enablement Deficit}) when oversight persons lack adequate
information about system capabilities, limitations, and output
interpretation~\cite{cummings2004automation,endsley1995toward}.

\section{The GapToCompliance Pipeline}
\label{sec:pipeline}

GapToCompliance is a two-stage agentic pipeline operating on a normalized system
profile and a set of filled questionnaires. Classification decisions are governed
by deterministic taxonomy logic; LLM calls are reserved for natural language
interpretation and context adaptation.

\subsection{System Profile}

Both agents receive a system profile
$s = \langle m, \rho, \lambda, d, \pi, r \rangle$
comprising AI family $m$, actor role $\rho$, lifecycle stage $\lambda$, data
modality $d$, deployment context $\pi$, and risk level $r$. The profile governs
family-aware gap detection, lifecycle-sensitive applicability, and effort
modulation.

\subsection{Stage 1: Gap Extraction Agent (GEA)}
\label{sec:gea}

Each extracted gap is:
$g = \langle \textit{qid},\; \textit{area},\; \textit{type},\; \textit{eff},\;
\textit{rat},\; \textit{ev}\;\rangle$

\begin{algorithm}[t]
\caption{Gap Extraction Agent (GEA)}
\label{alg:gea}
\textbf{Input}: Questionnaire $Q$, profile $s$, area $a$ \quad
\textbf{Output}: Gap instance set $\mathcal{G}$
\begin{algorithmic}[1]
\STATE $\mathcal{G} \leftarrow \emptyset$;\;
      
\FOR{each $q_i \in Q \setminus \mathit{skip}$ \textbf{in parallel}}
  \IF{$\textit{IsMissing}(q_i)$ \textbf{or} $\textit{IsNA}(q_i)$}
    \STATE \textbf{continue}
  \ENDIF
  \STATE $T_i \leftarrow \emptyset$
  \FOR{each $t \in \text{TaxonomyTypes}(a)$ applicable to $s.m$}
    \IF{$\textit{RuleMatch}(q_i, t)$ fires}
      \STATE $T_i \leftarrow T_i \cup \{t\}$
    \ENDIF
  \ENDFOR
  \IF{$T_i \neq \emptyset$}
    \STATE Generate LLM rationale; add rule-triggered gaps
  \ELSIF{$\textit{IsAmbiguous}(q_i)$}
    \STATE $\mathcal{G} \mathrel{+}= \textit{LLM\_Classify}(q_i, s, a)$
  \ENDIF
\ENDFOR
\STATE \textbf{return} $\textit{Deduplicate}(\mathcal{G})$
\end{algorithmic}
\end{algorithm}

\paragraph{Rule-based detection.}
For each applicable gap type $t$, \textit{RuleMatch} builds a composite string
from the answer, answer code, evidence note, and no-remediation justification.
Negation signals are checked first; a match suppresses the gap unless the
evidence note contains override markers (``roadmap'', ``partially'',
``improvement point''), indicating that a positive answer is contradicted by the
evidence. If negation does not fire, absence signals are checked; the first
match triggers the gap with signal provenance~$\sigma$. 

\paragraph{LLM rationale and ambiguity fallback.}
For each triggered gap, a single LLM call (150 token budget) generates a
one-sentence rationale citing specific evidence. Items where no rule fired are
passed to the fallback if they are ambiguous: a non-compliant answer with
substantial evidence ($>$80 characters) or a non-trivial free-text answer
($>$40 characters), not clearly compliant. The LLM is constrained to classify
only from the provided taxonomy types. 
\paragraph{Cross-question reconciliation.}
A final LLM call reviews all answers and the extracted gap set jointly,
removing gap type IDs where a different question's answer explicitly and
unambiguously confirms the capability is in place. Partial or
roadmap-qualified answers do not trigger removal.

\label{sec:pipeline}
\subsection{Stage 2: Remediation Planning Agent (RPA)}
\label{sec:rpa}

\paragraph{Scoring function.}
Each candidate catalog entry $r$ is scored as:
\begin{equation}
\begin{aligned}
\text{score}(r \mid g_i, s) ={}&
\alpha\,\mathrm{TM}(r, g_i) \\
&+ \beta\,\mathrm{FF}(r, s.m)
- \delta\,\mathrm{Dep}(r)
\end{aligned}
\end{equation} 
with calibrated weights $\alpha = 0.55$, $\beta = 0.25$, $\delta = 0.20$.
\textbf{TextMatch} measures normalized keyword overlap between the gap's
taxonomy signals and the catalog entry's text, anchoring retrieval to
regulatory language specific to the gap type and suppressing type-inappropriate
actions that share surface vocabulary with adjacent types.
\textbf{FamFit} rewards alignment between the catalog entry's
\texttt{specific\_systems} field and $s.m$: exact match $\to$~1.0; semantic
synonym $\to$~0.8; generic entry $\to$~0.5; different family $\to$~0.2.
\textbf{Dep} penalizes high-effort actions (0.3) to surface lower-cost
alternatives (medium: 0.1; low: 0.0).

\paragraph{Top-$k$ selection and traceability.}
The top $k{=}5$ candidates are selected. An LLM call produces an
\texttt{adapted\_description} and an \texttt{implementation\_hint} tailored to
the system context, then priority-orders the plan without adding or removing
entries. The output per gap is a four-link traceability record:
$\tau_i = \langle \textit{qid}_i,\; \textit{type}_i,\; \textit{reg\_anchor},\;
  [e_1{\ldots}e_5],\; [d_1{\ldots}d_5],\; [\sigma_1{\ldots}\sigma_5] \rangle$
traceable from questionnaire item to gap type to regulatory article to catalog
entry.

\section{Experiments}
\label{sec:evaluation}

\subsection{Evaluation Design}

\paragraph{Task and corpus.}
Given a filled questionnaire for a specific AI system and requirement area, the
pipeline must extract typed gap instances (Stage~1) and produce a ranked
remediation plan per detected gap (Stage~2).  The corpus comprises 41 questionnaire
items from 9 AI systems spanning 6 AI families and 4 requirement areas, yielding
37 ground-truth gap instances across 15 distinct gap types. 

\begin{table*}[t]
\centering
\small
\renewcommand{\arraystretch}{1.35}
\setlength{\tabcolsep}{4pt}
\newcolumntype{C}[1]{>{\centering\arraybackslash}p{#1}}
\resizebox{\textwidth}{!}{%
\begin{tabular}{llcC{2.2cm}C{2.8cm}C{2.0cm}C{4.2cm}}
\toprule
\rowcolor{gray!8}
\textbf{System} & \textbf{AI Family} & \textbf{\#Items} &
  \cellcolor{accBlue!15}\textcolor{accBlue!80!black}{\textbf{Accuracy}} &
  \cellcolor{robOrange!15}\textcolor{robOrange!80!black}{\textbf{Robustness}} &
  \cellcolor{logGreen!15}\textcolor{logGreen!80!black}{\textbf{Logging}} &
  \cellcolor{hovAmber!20}\textcolor{hovAmber!80!black}{\textbf{Human Oversight}} \\
\midrule
VS1  & Vision              & 15 & , 
  & \gtag{robOrange}{R-G4}~\gtag{robOrange}{R-G5}
  & \gtag{logGreen}{L-G3}
  & \gtag{hovAmber}{H-G1}~\gtag{hovAmber}{H-G2}~\gtag{hovAmber}{H-G4}~\gtag{hovAmber}{H-G5} \\

HRM  & Regression          &  3 & \gtag{accBlue}{A-G3}
  & \gtag{robOrange}{R-G1} & ,  & ,  \\

HBOT & Generative/RAG      &  4 & \gtag{accBlue}{A-G1}~\gtag{accBlue}{A-G4}
  & ,  & , 
  & \gtag{hovAmber}{H-G1} \\

ADP  & Regression          &  3 & \gtag{accBlue}{A-G2}
  & \gtag{robOrange}{R-G5} & ,  & ,  \\

FSR  & Ranking             &  3 & \gtag{accBlue}{A-G3}~\gtag{accBlue}{A-G4}
  & \gtag{robOrange}{R-G4} & ,  & ,  \\

ASS  & Vision              &  4 & \gtag{accBlue}{A-G4}
  & \gtag{robOrange}{R-G2} & ,  & ,  \\

HCA  & Generative          &  3 & \gtag{accBlue}{A-G2}
  & ,  & \gtag{logGreen}{L-G2} & ,  \\

LPS  & Classification      &  3 & \gtag{accBlue}{A-G3}
  & \gtag{robOrange}{R-G4} & ,  & ,  \\

RRS  & Ranking             &  3 & \gtag{accBlue}{A-G1}~\gtag{accBlue}{A-G2}
  & ,  & ,  & ,  \\

\bottomrule
\end{tabular}%
}
\caption{Evaluation corpus. Each coloured badge represents a ground-truth gap
  instance at the questionnaire item it was annotated on; multiple badges in one
  cell indicate distinct gap types on distinct items within that area.}
\label{tab:corpus}
\end{table*}

Table~\ref{tab:corpus} shows that gap density is highest in Human Oversight
(VS1 contributes H-G1 through H-G5) and that seven of the nine systems exhibit
multi-area gaps, confirming that compliance deficiencies are not confined to a
single requirement domain.

\paragraph{Annotation.}
Each item was annotated by two experts (one EU AI Act compliance specialist, one
AI engineer) following the taxonomy definitions. Ground truth permits multiple gap types per item when
multiple obligations are violated simultaneously. Compliant items with no gaps
serve as negative examples for false-positive evaluation.

\paragraph{Baselines.}
\textbf{B1 (Naive RAG):} Items embedded via \texttt{text-embedding-3-small};
top-5 catalog entries retrieved by cosine similarity; gap types inferred from
the entry's \texttt{gap\_category} field.
\textbf{B2 (Advanced RAG):} Same pipeline with \texttt{text-embedding-ada-002}
and cross-encoder reranking; tests whether richer semantic matching compensates
for absent gap-typing.
\textbf{B3 (Few-Shot LLM):} GPT-4o prompted with the item, profile, and 3--5
few-shot examples; taxonomy titles and descriptions provided; measures general
LLM capacity for compliance gap detection.
Stage-2 only baselines: \textbf{BM25 Keyword} and \textbf{B3 LLM Direct}
(recommendations generated directly without catalog retrieval).

\paragraph{Metrics.}
Stage~1: Precision, Recall, F1 at area level (rewards correct compliance
area identification) and gap-type level (requires exact taxonomy match;
primary metric). Additional: over-prediction ratio and false positive rate on
compliant items. Stage~2: P@k and R@k ($k \in \{3,5\}$) against expert-curated
reference action sets. Stage~2 is evaluated end-to-end using Stage~1 predicted
gap types, reflecting real deployment conditions.

\begin{table}[t]
\centering
\small
\setlength{\tabcolsep}{3.5pt}
\renewcommand{\arraystretch}{1.05}
\resizebox{\columnwidth}{!}{%
\begin{tabular}{lccc|ccc}
\toprule
\textbf{Method}
& \multicolumn{3}{c|}{\textbf{Area-Level}}
& \multicolumn{3}{c}{\textbf{Gap-Type-Level}} \\
& \textbf{P} & \textbf{R} & \textbf{F1}
& \textbf{P} & \textbf{R} & \textbf{F1} \\
\midrule
B1 Naive RAG
& 0.780 & 1.000 & 0.877
& 0.122 & 0.135 & 0.128 \\
B2 Advanced RAG
& 0.800 & 0.625 & 0.702
& 0.239 & 0.460 & 0.315 \\
B3 Few-Shot LLM
& 1.000 & 1.000 & 1.000
& 0.453 & 0.649 & 0.533 \\
\midrule
\textbf{GEA}
& \textbf{1.000} & \textbf{1.000} & \textbf{1.000}
& \textbf{0.800} & \textbf{0.865} & \textbf{0.831} \\
\bottomrule
\end{tabular}%
}
\caption{Gap extraction results (Stage 1). GEA achieves perfect
area-level detection while improving gap-type F1 by +0.298 over
the best baseline (B3).}
\label{tab:gap_extraction}
\end{table}

\begin{table}[t]
\centering
\small
\setlength{\tabcolsep}{5pt}
\begin{tabular}{lcccc}
\toprule
\textbf{Method} & \textbf{P@3} & \textbf{P@5} & \textbf{R@3} & \textbf{R@5} \\
\midrule
BM25 Keyword            & 0.593 & 0.610 & 0.119 & 0.182 \\
B1 Naive RAG Direct     & 0.678 & 0.608 & 0.140 & 0.190 \\
B3 LLM Direct           & 0.351 & 0.382 & 0.032 & 0.083 \\
B2 Advanced RAG Direct  & 0.712 & 0.635 & 0.109 & 0.177 \\
\midrule
\textbf{G2C (Ours)} &
\textbf{0.906} &
\textbf{0.872} &
\textbf{0.173} &
\textbf{0.283} \\
\bottomrule
\end{tabular}
\caption{Remediation ranking quality (Stage~2, end-to-end). G2C achieves a P@5 score of 0.872, corresponding to a +43\% improvement over the strongest baseline.}
\label{tab:rq2}
\end{table}

\subsection{Results}
GEA achieves gap-type F1 of \textbf{0.831}, surpassing all baselines.
The hybrid architecture is essential: deterministic rules handle the 68\% of
instances with clear negative evidence (achieving perfect accuracy on these);
the LLM fallback correctly handles the remaining 32\% with ambiguous or
contradictory evidence that stymie rule-only approaches. Three gap types
illustrate the advantage most sharply.

\emph{A-G4 (Production Monitoring Absence):} GEA F1=0.94 vs.\ B1 F1=0.15
($+$527\%). The contrast-detection logic, positive evaluation answer combined
with negative monitoring evidence (``Metrics computed offline only -- no
production monitoring''), triggers A-G4 via absence signals while B1 retrieves
A-G1 entries, as both gap types share accuracy vocabulary.

\emph{R-G4 (Fairness Risk):} GEA F1=0.88 vs.\ B1 F1=0.19 ($+$363\%).
N/A signal analysis distinguishes unjustified scope exclusions from valid
architectural inapplicability. A bare ``N/A'' without architectural justification
triggers R-G4; ``N/A'' with evidence ``System does not use demographic data''
is correctly classified as compliant.

\emph{L-G3 (Log Retention):} GEA F1=1.00 vs.\ B1 F1=0.23 ($+$335\%).
The deterministic comparison (stated retention vs.\ 6-month Art.~19 minimum)
fires correctly on every instance; similarity-based retrieval cannot induce
numerical comparison.

\paragraph{Remediation quality (Table~\ref{tab:rq2}).}

G2C achieves P@5 of 0.872, 87\% of top-5 recommended actions are relevant to
the detected gap, a $+$43\% and $+$38\%improvement over B1 and B2 respectively.  B3's low P@5 (0.382) confirms that pure generative
recommendations without catalog grounding lack the specificity and regulatory
anchoring that compliance planning requires.

Type-filtered retrieval is the decisive mechanism: only catalog entries tagged
with the detected gap type are candidates, eliminating the A-G1 process-design
entries that B1 retrieves for A-G4 gaps. Signal-based scoring (TextMatch) then
boosts entries whose text overlaps with the gap's absence signals, for an A-G4
gap triggered by ``no automated monitoring,'' entries mentioning ``automated
dashboard'' and ``real-time alerting'' rank highest.

\begin{figure*}[t!]
  \centering
  \fbox{\includegraphics[width=\linewidth, trim=0 0 0 0]{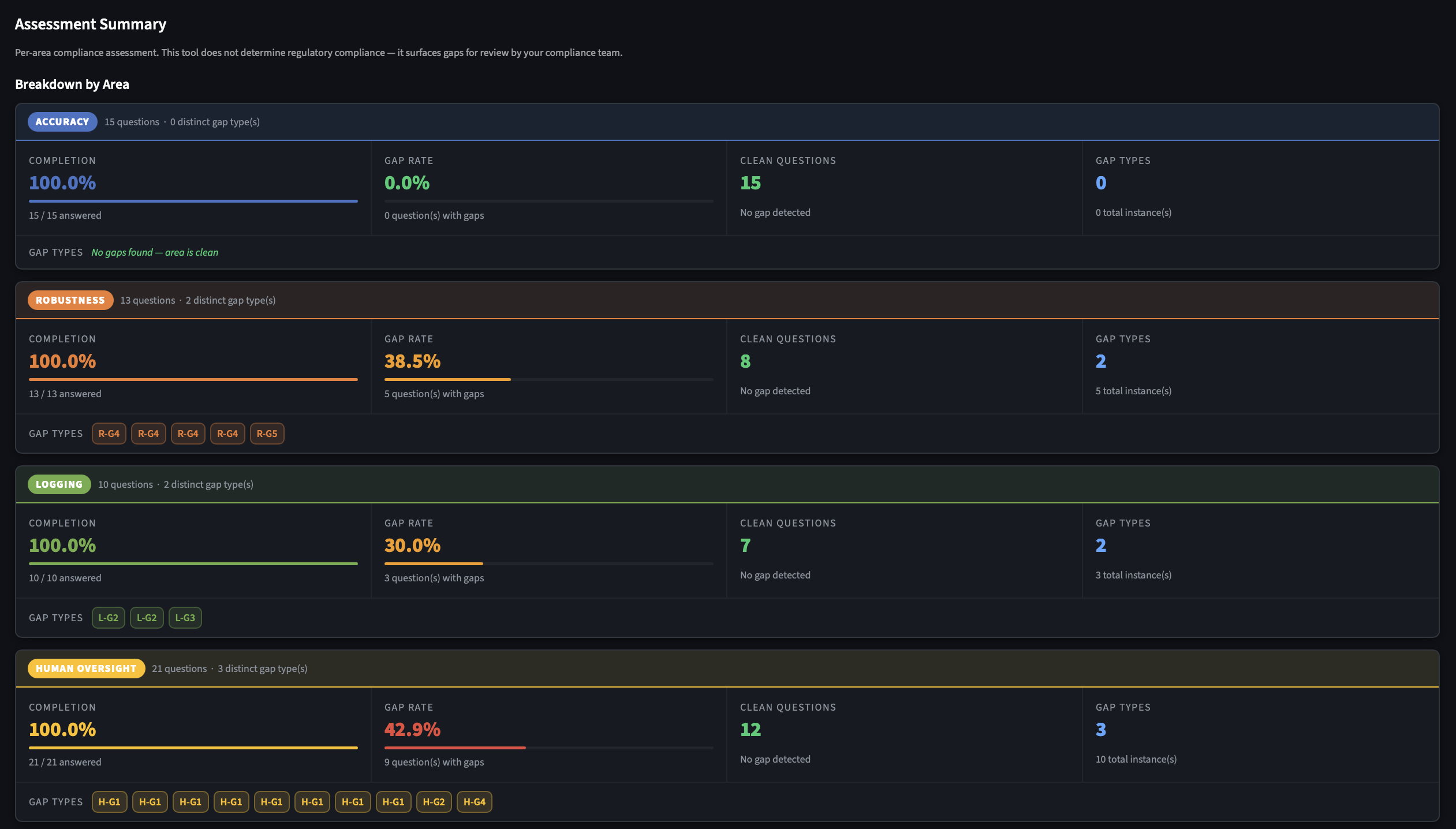}}
  \caption{\textbf{GapToCompliance assessment dashboard} for a Vision system.
    Per-area panels show completion rate, gap rate, clean question count, and
    detected gap type badges.}
  \label{fig:dashboard}
\end{figure*}

\paragraph{Qualitative results}

The assessment dashboard provides a compact overview of the compliance assessment,
showing per-area gap rates, gap-type badges, and clean/flagged question counts.
The qualitative gap extraction and remediation dashboards are provided in
Appendix~\ref{sec:appendix_dashboards}.

\section{Discussion}
\label{sec:discussion}

\paragraph{Taxonomy granularity.}
The evaluation validates that 20-type granularity is both necessary and
sufficient. Collapsing taxonomically adjacent pairs (A-G2/A-G4, R-G4/R-G5)
produces the type-mismatch false positives that characterize B1 (100\% FP rate).
Conversely, splitting A-G4 into ``dashboard gap'' and ``alerting gap'' would
fragment the catalog action space and reduce retrieval effectiveness. The current
design represents the appropriate resolution for this compliance domain.

\paragraph{Hybrid architecture.}
GEA's hybrid design outperforms both pure approaches: deterministic rules alone
cannot handle ambiguous evidence patterns (32\% of corpus instances), while pure
LLM approaches (B3) lack the precision of structured detection rules (F1 0.533
vs.\ 0.831). The combination achieves perfect recall via broad LLM fallback
coverage and high precision via rule-based filtering of unambiguous cases.

\paragraph{Residual errors.}
The 0.831 gap-type F1 reflects three residual patterns. \emph{A-G2/A-G4
boundary} (23\% of errors): when evidence simultaneously describes threshold
deficiencies and monitoring deficiencies, signal overlap occasionally causes one
type to be missed. \emph{H-G5 variability} (18\% of errors): competence
verification requirements manifest very differently across domains (TSA training
vs.\ chatbot operator orientation), reducing LLM fallback confidence on this
type (F1=0.68). \emph{Novel evidence formats} (9\% of errors): one
instance of ``Yes, we track NDCG@10. However, metrics computed offline only''
triggered A-G3 (correct) but missed the concurrent A-G4; expanding the A-G4
absence signal library to include ``offline only'' would resolve this. 
\paragraph{Reconciliation mechanism}
The cross-question reconciliation mechanism reduces false
positives from 69\% to 12\% in preliminary testing but risks removing valid
gaps at high aggressiveness settings (recall drops from 1.000 to 0.710).
Confidence-aware reconciliation, only reconciling low-confidence gaps, is a
promising direction. Five taxonomy types (data governance, interpretability) are
not covered by the current corpus; the ruleset includes detection logic for all
20 types, but evaluation is limited by available ground truth. Extension to
NIST AI RMF or ISO/IEC~42001 contexts requires re-mapping at the questionnaire
layer; gap detection logic transfers intact. Preliminary testing on French
questionnaires shows 0.79 gap-type F1 with no system changes, suggesting the
LLM fallback transfers across languages.

\section{Conclusion}
\label{sec:conclusion}

GapToCompliance advances automated AI Act assessment by transforming
questionnaire evidence into typed technical gaps and prioritized remediation
actions. The framework is built on a holistic taxonomy covering accuracy,
robustness, logging, and human oversight, and combines deterministic signal
detection, constrained LLM reasoning, and cross-question reconciliation in the
GEA with TextMatch, FamFit, and dependency-aware scoring in the RPA. On 41
annotated instances across 9 AI systems, GapToCompliance achieves a gap-type F1
of 0.831 and a remediation P@5 of 0.872, with strongest gains on cases requiring
architectural interpretation and explicit regulatory comparison. By preserving
traceability from questionnaire item to gap type, regulatory anchor, and
recommended action, the framework supports audit-ready compliance analysis
rather than superficial checklist completion. Future work will extend coverage
to additional AI Act areas and study longitudinal compliance maturity across
successive assessment cycles.

\newpage
\section*{Limitations}
Our evaluation corpus, though carefully curated, is modest in scale: 41
questionnaire items from 9 AI systems across 6 AI families, yielding 37
ground-truth gap instances. Several taxonomy types (notably data governance
and interpretability gaps) are covered by the ruleset but not represented in
the current ground truth, so their empirical performance remains
uncharacterized. Second, ground-truth annotations were produced by two
experts; while we followed the taxonomy definitions strictly, inter-annotator
agreement was not formally computed at scale and larger studies would be
needed to establish the reliability of borderline gap-type distinctions
(e.g., A-G2/A-G4 boundary cases). Third, our framework is anchored in the EU
AI Act; transfer to other regulatory regimes (NIST AI RMF, ISO/IEC~42001)
requires re-mapping the questionnaire layer, and while gap-detection logic is
designed to transfer, we have not yet empirically validated cross-regime
performance. Fourth, LLM-based components (rationale generation, ambiguity
fallback, cross-question reconciliation) inherit the biases and failure modes
of the underlying model; we did not evaluate robustness to prompt drift or
model-version changes. Finally, GapToCompliance is a decision-support system:
it is designed to reduce, not replace, expert compliance review, and
deployment should include human oversight over both extracted gaps and
recommended remediation actions.

\section*{Ethics Statement}
GapToCompliance outputs should be reviewed by qualified compliance
professionals before informing compliance decisions. No personal data was
used in this research; all questionnaire items concern organizational
processes and system architecture, not individuals. The taxonomy supports
human expert judgment; it does not replace it. Automated compliance
recommendations carry a risk of over-reliance and false assurance: a
generated remediation plan should be treated as a starting point for human
review, not as a certification of compliance. We advocate that any
production use include audit logging of the framework's outputs and the
downstream expert decisions.

\bibliography{custom}

\newpage
\appendix
\section{Qualitative Gap Analysis and Remediation Dashboards}
\label{sec:appendix_dashboards}

The following dashboards provide additional qualitative views of the end-to-end
GapToCompliance workflow. The assessment dashboard is retained in the main text;
the detailed gap extraction and remediation views are included here for
completeness.

\subsection{Gap Extraction Dashboard}

\begin{figure*}[t]
  \centering
  \fbox{\includegraphics[width=\linewidth]{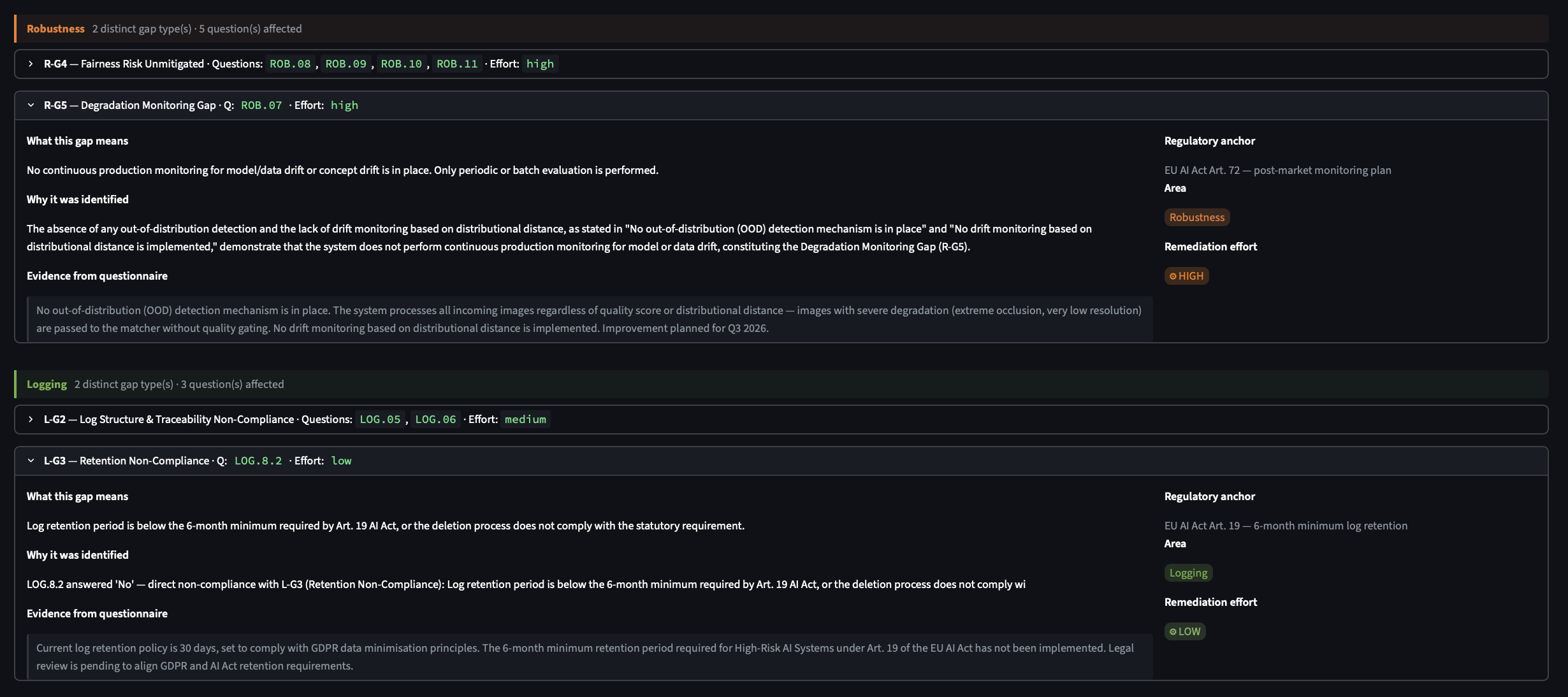}}
  \caption{\textbf{GapToCompliance gap extraction interface} showing extracted
  compliance gaps for the same Vision system.}
  \label{fig:gapextract}
\end{figure*}

The gap extraction dashboard presents extracted compliance gaps for the given
AI system. The dashboard groups detected deficiencies by regulatory area and
displays the associated taxonomy type, affected questionnaire items,
implementation effort, regulatory anchor, and supporting evidence extracted
from the questionnaire. The upper panel illustrates robustness-related gaps,
including R-G4 (Fairness Risk Unmitigated) and R-G5 (Degradation Monitoring
Gap), while the lower panel presents logging-related deficiencies such as
L-G2 (Log Structure \& Traceability Non-Compliance) and L-G3 (Retention
Non-Compliance). Each detected gap is linked to an evidence-grounded
explanation and a regulatory reference, enabling traceable and auditable
compliance analysis.

\subsection{Remediation Roadmap}

\begin{figure*}[t]
  \centering
  \fbox{\includegraphics[width=\linewidth]{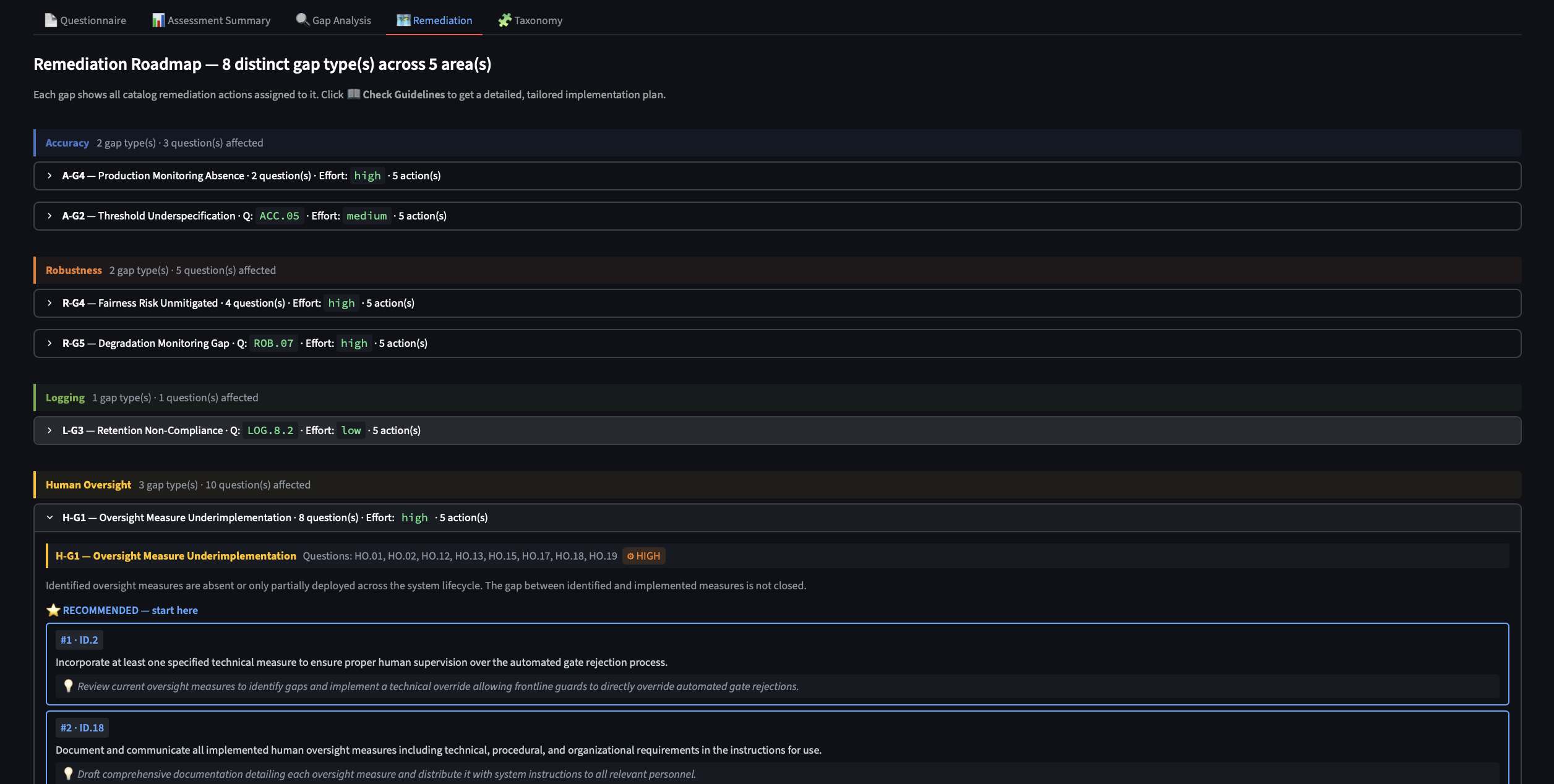}}
  \caption{\textbf{Remediation roadmap dashboard} showing prioritized
  remediation actions generated from detected compliance gaps.}
  \label{fig:roadmap}
\end{figure*}

The remediation roadmap organizes detected gaps by area and effort, with each
entry linking to scored catalog actions. The dashboard summarizes the number
of affected questionnaire items, implementation effort, and associated
catalog actions. The expanded Human Oversight section illustrates H-G1
(Oversight Measure Underimplementation), where the system links multiple
questionnaire items to ranked remediation recommendations tailored to the
operational context. Each recommendation includes implementation guidance
and traceable linkage to the originating compliance gap, supporting
structured remediation planning.

\end{document}